\documentclass{bmvc2k}

\usepackage{graphicx}
\usepackage{epsfig}
\usepackage{graphicx}
\usepackage{amsmath}
\usepackage{amssymb}
\usepackage{amsthm}
\usepackage{mathrsfs}
\usepackage{algorithm}
\usepackage{algpseudocode}
\usepackage{graphics}
\usepackage{booktabs}
\usepackage{multirow}
\usepackage{pifont}
\usepackage{xcolor}
\usepackage{ulem}    

\newcommand{\cmark}{\textcolor{black!50!white}{\ding{51}}}
\newcommand{\xmark}{\textcolor{black!50!white}{\ding{55}}}

\title{ Back to the Future: Cycle Encoding Prediction for Self-supervised Video Representation Learning\vspace{-8pt}}

\addauthor{Xinyu Yang}{xinyu.yang@bristol.ac.uk}{1}
\addauthor{Majid Mirmehdi}{m.mirmehdi@bristol.ac.uk}{2}
\addauthor{Tilo Burghardt}{tilo@cs.bris.ac.uk}{3}

\addinstitution{
Department of Computer Science\\
Faculty of Engineering\\
University of Bristol\\
Bristol, UK \\
}

\runninghead{Yang\textit{ et al.}}{Back to the Future....}

\begin{document}

\maketitle
\vspace{-22pt}
\begin{abstract}
We show that learning video feature spaces in which temporal cycles are maximally predictable benefits action classification. In particular, we propose a novel learning approach, Cycle Encoding Prediction~(CEP), that is able to effectively represent high-level spatio-temporal structure of unlabelled video content. CEP builds a latent space wherein the concept of closed forward-backward, as well as backward-forward, temporal loops is approximately preserved. As a self-supervision signal, CEP leverages the bi-directional temporal coherence of entire video snippets and applies loss functions that encourage both temporal cycle closure and contrastive feature separation. Architecturally, the underpinning network architecture utilises a single feature encoder for all input videos, adding two predictive modules that learn temporal forward and backward transitions. We apply our framework for pretext training of networks for action recognition and report significantly improved results for the standard datasets UCF101 and HMDB51.
\end{abstract}


\vspace{-16pt}
\section{Introduction}
\vspace{-6pt}
\label{sec:intro}
Videos constitute highly structured objects which offer rich and intrinsically correlated information that is often suitable as a self-supervision signal for unsupervised representation learning. Yet, defining which exact aspects of the video should be exploited for effective learning of semantically relevant embeddings, and how the resulting latent spaces are to be constructed, structured, and constrained, is a topic of {active research~\cite{bengio2013representation,doersch2015unsupervised,feichtenhofer2016convolutional,carreira2017quo,wang2018non,feichtenhofer2019slowfast,feichtenhofer2020x3d}.}

\begin{figure}[ht]
	\centering
	\includegraphics[width=300pt]{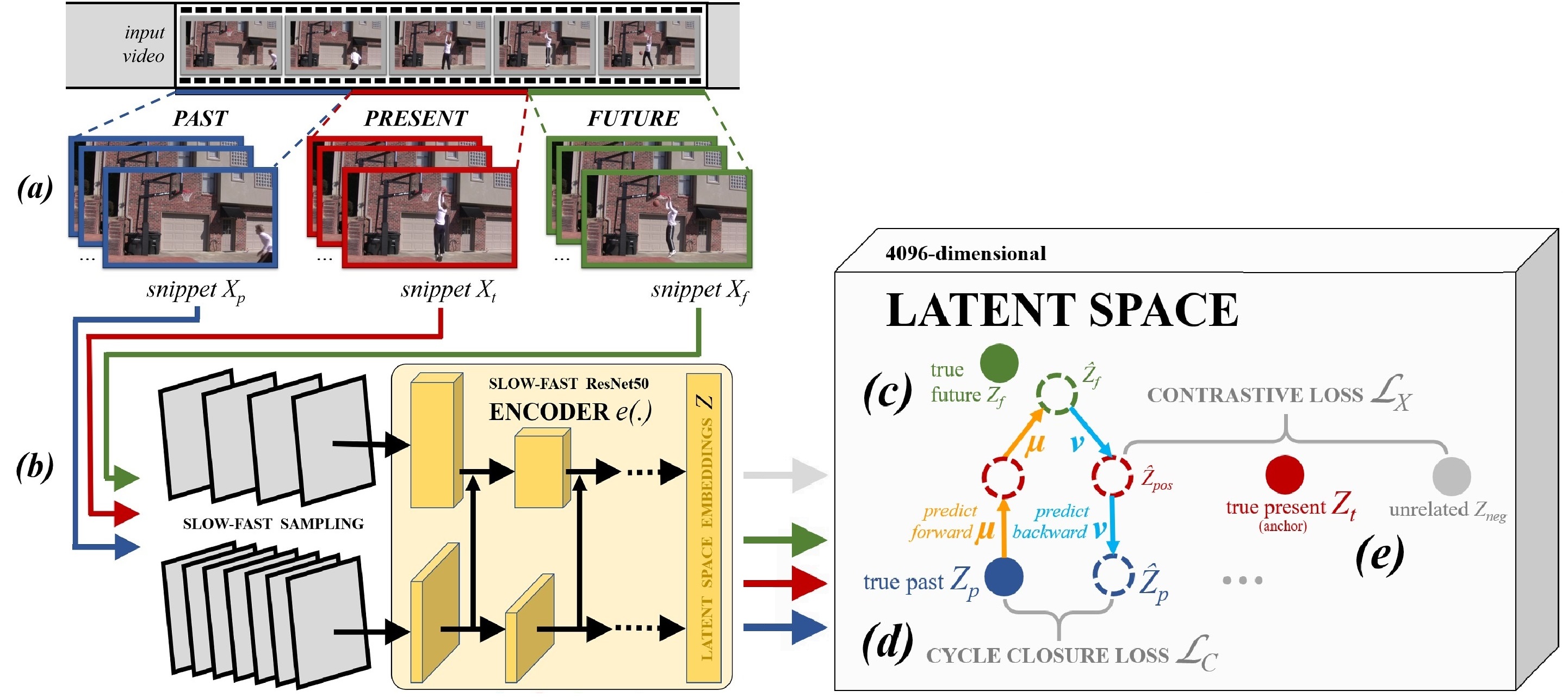}\vspace{-6pt}
	\caption{\footnotesize{\textbf{Overview of Self-Supervised CEP Approach.} We represent videos in a feature space wherein bi-directional temporal cycles are maximally predictable. \textbf{\textit{(a)}}~The video data is sampled into snippets where each snippet~$X_t$ has a past~$X_p$ and future~$X_f$ neighbour. \textbf{\textit{(b)}}~In pretext learning, a latent space~$Z_t=e(X_t)$ is generated via an encoder~$e$ {(such as 3DResNet or SlowFast)}. \textbf{\textit{(c)}}~In this space, we encourage that learnt temporal forward~$\mu$ and backward~$\upsilon$ predictions form closed cycles (example cycle from past~$Z_p$ to some future~$\hat{Z}_f$ and back to some past~$\hat{Z}_p$ shown). \textbf{\textit{(d)}}~Embedding~$e$ and predictors~$\mu$ and $\upsilon$ are co-trained in  self-supervised mode, minimising a loss~$\mathcal{L}_C$ that favours cycle closure. \textbf{\textit{(e)}}~Further, we apply a contrastive loss $\mathcal{L}_{X}$ {in concert to} encourage feature separation.}}
	\label{fig:overview}
	\vspace{-15pt}
\end{figure}

In this paper, we take a closer look at exploiting the bidirectional temporal structure of video to support the learning of semantically relevant high-level spatio-temporal feature spaces.
In particular, as illustrated in Fig.~\ref{fig:overview}, we show that learning feature spaces, in which temporal cycles (going back and forth in time) become maximally predictable, can benefit action classification. Intuitively, any latent space that is suitable for classifying actions should carry a transitive structure which encodes the distinctive 'playing out' of actions. Walking along a path  of forward predictions followed by an equal number of backward predictions (a bi-directional temporal cycle) should cancel out and provide an exploitable self-supervision criterion for pre-training. Note, this does not model multi-hypothesis predictions~\cite{suris2021learning,vondrick2016anticipating}.

Following this idea, we propose a novel pretext learning approach, named Cycle Encoding Prediction~(CEP), which operates on unlabelled video and is driven by bi-directional temporal cycle closure. It aims at constructing a latent embedding wherein sequences of temporal forward-backward and backward-forward predictions form approximately closed cycles. Meanwhile, similar video states are, as per contrastive learning, still mapped to nearby locations. Thus, to generate this space, CEP enforces loss functions that deliver both bi-directional temporal cycle closure and contrastive feature separation. To cope with GPU memory restrictions, we use a memory bank to generate negative features for our contrastive learning. Further, to avoid trivial solutions that are  non-semantic-bearing, we introduce Synchronized Temporal-Group Normalization as well as exploit clues from optical flow. The proposed CEP approach is fundamentally different from future-frame prediction works such as  DPC \cite{han2019video}, TCC \cite{dwibedi2019temporal}, and  TimeCycle \cite{wang2019learning}. For example, DPC \cite{han2019video} is a {\it uni-directional} predictive contrastive model that aggregates temporal embeddings recurrently to make predictions,
and in TimeCycle \cite{wang2019learning}, patch features from a sequence of frames, generated by tracking forward and backward to learn the affinity matrix between nearby frames. In contrast, CEP jointly learns high level temporal cycle structure via a closure loss and the prediction feature space itself via a contrastive loss (see Fig.~\ref{fig:overview}).

We note that the fundamental differences between our work and the main body of work on contrastive cycle consistency, such as  \cite{han2019video,wang2019learning, Jabri2020, Gordon2020}, are that we employ temporal cycle closure as a direct loss objective in a \textit{latent semantic space of video snippets} -- thus, CEP is not bound to single frame association, appearance-similarity or tracking tasks. Instead, CEP provides temporally guided, \textit{semantic} self-supervision that encodes entire video segments and their temporal relationships, e.g. to further action recognition.

Our key contributions are: (i)~we present a novel temporal cycle-exploiting minimisation objective for self-supervised pretext learning that provides superior generalisation and representational ability in latent space, (ii)~we introduce the Cycle Encoding Prediction  approach that implements this minimisation objective in combination with contrastive learning and by adding key components, such as a memory bank and ways to avoid trivial solutions, (iii)~we quantitatively compare different self-supervised video representation learning architectures, while different design choices are also investigated to consider the contributions of each component,  (iv) we achieve competitive or better than the state-of-the-art results (depending on the pretext training dataset, architecture, etc.) performance on standard datasets UCF101 and HMDB51. Our source code is available at \url{https://github.com/youshyee/CEP}.
\vspace{-15pt}
\section{Related Work}
\vspace{-5pt}


{\bf Spatial Representation Learning --} Single image frames already provide a wealth of intrinsic spatial and colour channel relationships which form structure and can be utilised for unsupervised learning of the content-specific structure. Tasks such as colourising grey-scale images~\cite{larsson2017colorization,zhang2016colorful}, image inpainting~\cite{pathak2016context}, predicting relative image patch positions~\cite{doersch2015unsupervised}, or image jigsaw puzzles~\cite{noroozi2016unsupervised,kim2018learning} leverage this information. However, whilst single frames can  capture the essence of many high-level semantic concepts, such as actions~\cite{simonyan2014two} (and self-supervision may be able to learn this information~\cite{wang2015unsupervised}), their expressiveness is limited with regard to many motion-dependent~\cite{martinez2017human} and fine-grained~\cite{mac2019learning} action recognition tasks.

{\bf Spatial-Temporal Representation Learning --} Videos contain rich spatio-temporal information that is naturally suitable for unsupervised learning.
Many existing works act on unlabelled videos for spatio-temporal representation learning, e.g.~\cite{wang2015unsupervised,misra2016shuffle,lee2017unsupervised,kim2019self,han2019video,yao2020video,fernando2017self,wang2019learning,uvc_2019,dwibedi2019temporal,feichtenhofer2021large,han2020self}. Temporal coherence and dynamics of video were exploited for self-supervised learning early in \cite{wang2015unsupervised}.
Positive samples from the tracker, together with the negatives that were sampled randomly, were deployed in a contrastive learning framework with a pre-designed Siamese-triplet network for self-supervised learning. Alternatively, one may leverage the temporal continuity of video snippets to design the proxy task~\cite{misra2016shuffle,lee2017unsupervised,xu2019self}. Shuffle \& Learn~\cite{misra2016shuffle} discriminates the non-chronologically ordered frames from a video to enforce the learning of temporal relations among frames. Similarly,
OPN~\cite{lee2017unsupervised} designs the learning task by reordering the shuffled frames from a video which enforces the learning of temporal continuity.
In VCOP~\cite{xu2019self}, by relying on the temporal coherence within snippets from a shuffled video, predicting the order becomes the proxy task for self-supervised learning.
Rather than sorting whole frames,
ST-puzzle~\cite{kim2019self} focuses on individual non-overlapped spatial patches to introduce a proxy task to re-order permuted 3D spatio-temporal  snippets.



Since the introduction of GANs~\cite{NIPS2014_5423}, their application towards constructive video generation and prediction have received a great deal of interest.
In~\cite{wang2019self}, video frames are partitioned into several spatial regions by different appearance and motion patterns. Exploiting the statistical relationship within the patterns for a known frame can then be used to derive relationships in unknown frames by a dual-stream network.      
DPC~\cite{han2019video} observes an earlier period of a video with a recurrent network for global context extraction. Then, the network makes a prediction for the context of a later period of the video using the Noise Contrastive Estimation (NCE) loss as the learning constraint. Mem-DPC~\cite{Han20memdpc} extends DPC by appending a memory mechanism that queries a similar pattern from memory when making predictions.

More recent works leverage speed information from videos to form a self-supervised learning task, e.g. \cite{yao2020video, benaim2020speednet,chen2020rspnet}. PRP~\cite{yao2020video} considers the semantic similarity and temporal structure difference of the same video at different playback rates, introducing  multi-task learning objectives that enforce the model to enrich the detail temporally 
as well as to regress the pre-defined playback rate.
 SpeedNet~\cite{benaim2020speednet} seeks to learn the appearance and motion from unlabelled videos by predicting the playback rate for the clip. RSPNet~\cite{chen2020rspnet} extends this to a relative speed perception task by applying speed augmentation strategy to make the network consider appearance related content. CVRL~\cite{qian2020spatiotemporal}, as the video extension of SimCLR~\cite{chen2020simple}, performs contrastive learning by augmenting video clips spatially and temporally.

Despite substantial progress in self-supervised learning by prediction, existing methods have not yet fully exploited the representational  potential of temporal predictions within video.
In this paper, we address this problem and describe the learning of video feature spaces where cycle-consistent bi-directional temporal predictions are optimised in tandem with contrastive feature separation.

\vspace{-10pt}
\section{Cycle Encoding Prediction}
\vspace{-5pt}
We would like to utilise unlabelled video and its property of temporal cycle closure to pretext learn a latent space for content encoding. To do this, we exploit the bi-directional temporal coherence of video as a self-supervision signal and enforce loss functions that encourage  temporal cycle closure, as well as contrastive feature separation.



{\bf Framework --} CEP operates on video snippets~$X_t \in \mathbb{R}^{T\times H \times W \times C} $
harvested from the video stream (as illustrated in Fig.~\ref{fig:overview}), where {{$T\times H \times W \times C$} are, time, height, width, and channel respectively}. Given a particular snippet~$X_t$ representing the current state, there are two adjacent neighbouring snippets: the past snippet~$X_p=X_{t-1}$, and the future snippet~$X_f=X_{t+1}$. We consider snippets being encoded within a latent space via a non-linear network~$e(\cdot)$, a function which we would like to learn and which should preserve temporal cycle closure and contrastive feature separation. Formally,~$e(\cdot)$ encodes input video snippets~$X_t$ as locations~$Z_t$ in latent space, \textit{i.e.} 	$Z_t=e(X_t)$,
where 
the latent space $Z_t$ is a feature of the form $\mathbb{R}^{D}$.

\begin{figure}[t]
	\centering
	\includegraphics[width=160pt,height=77pt]{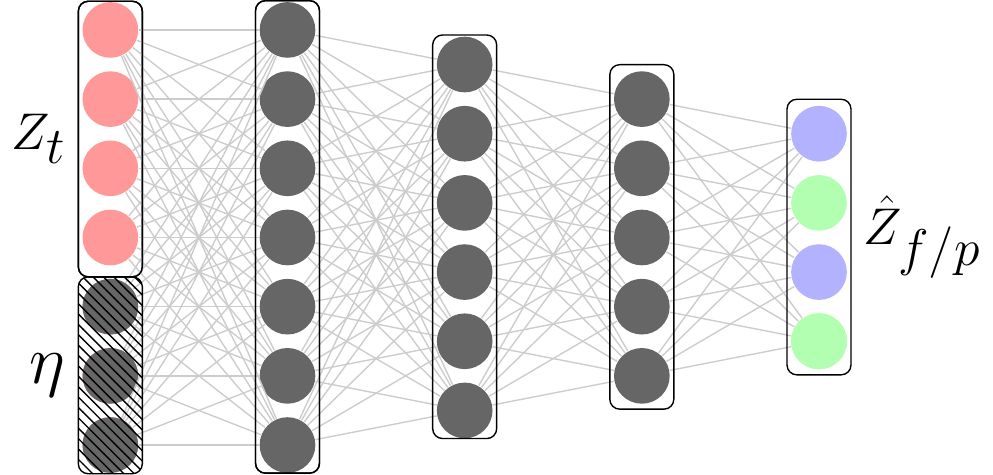}\hspace{20pt}
	\includegraphics[width=153pt,height=90pt]{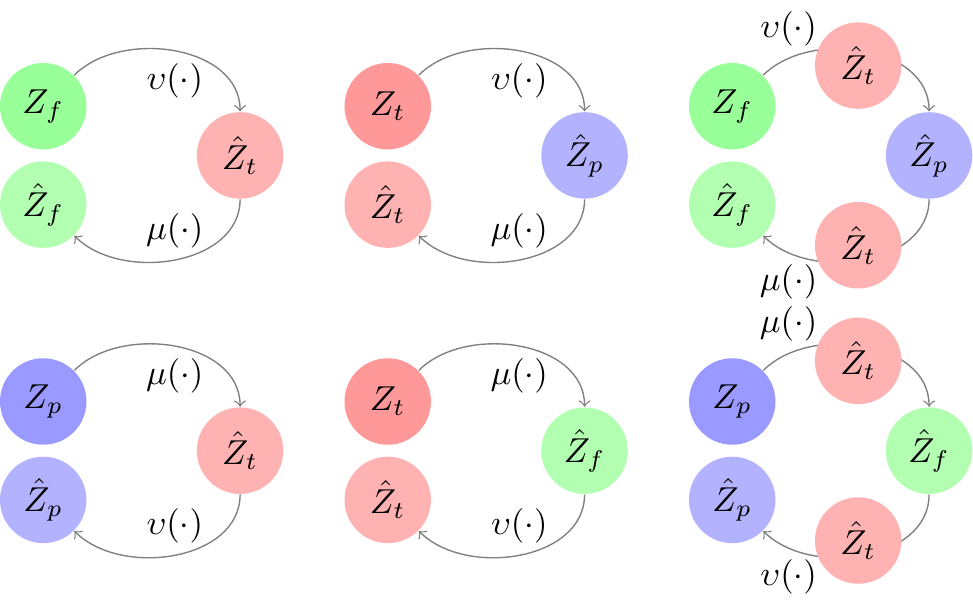}
	\vspace{-4pt}
	\caption{\footnotesize{\textbf{\textit{(left)} Structure of Predictors $\mu$ and $\upsilon$}. An embedding $Z_t$ concatenates with white noise $\eta$ as the input. The two predictors are implemented with 4-layers MLP with ReLu activation after each layer. Note that future prediction function and past prediction function share this same architecture, but they do not share parameters in training. \textbf{\textit{(right)} Elementary Cycles in Embedding Space.} The current, future, and past states are {represented} in pink, green, and purple respectively. The depicted 6 most fundamental cycles are considered for cycle consistency. Note that apart from these elementary cycles, the concept could be extended and longer loops are possible, but they require increasing computational resource and a related analysis would be beyond the scope of this paper.}}\label{fig:predictors}
	\vspace{-12pt}
\end{figure}

On the most fundamental level, we note that if the future state $Z_f=e(X_f)$ or past state $Z_p=e(X_p)$  can be predicted by current $Z_t$, then the latent space representation must have preserved elementary aspects of the temporal structure of input video clips. To implement this concept, we introduce two predictive functions~$\mu(\cdot, \eta)$ and $\upsilon(\cdot, \eta)$ to predict future and past states, respectively. We add white noise $\eta$ in both predictive functions to increase the stochasticity. As exemplified in Fig.~\ref{fig:predictors}~\textit{(left)}, $\mu(\cdot)$ can take current state~$Z_t$ and past state~$Z_p$, and predict the future by the current and the current by the past, such that\vspace{-6pt}
\begin{equation}
	\hat{Z}_f=\mu(Z_t,\eta)=\mu(e(X_t),\eta)  \mbox{~~~and~~~}
	\hat{Z}_t=\mu(Z_p,\eta)=\mu(e(X_p),\eta).\vspace{-6pt}
\end{equation}
Similarly, the elementary function~$\upsilon(\cdot)$ allows for the following direct predictions, \textit{i.e.},\vspace{-6pt}
\begin{equation}
	\hat{Z}_p=\upsilon(Z_t,\eta)=\upsilon(e(X_t),\eta)   \mbox{~~~and~~~}
	\hat{Z}_t=\upsilon(Z_f,\eta)=\upsilon(e(X_f),\eta).\vspace{-6pt}
\end{equation}
Note that we clearly differentiate between latent space locations~$Z_t$ arising directly from the embedding function and~$\hat{Z}_t$ arising from predictions solely within latent space.

{\bf Consistency of Predicted Cycles --} We observe that the predictive functions can be applied to embeddings recursively, as depicted in Fig.~\ref{fig:predictors}~\textit{(right)}. Such application allows for the generation of entire cycles of predictions knowing that the future operator~$\mu(\cdot)$ is the reverse of the past operator~$\upsilon(\cdot)$, \textit{i.e.} $\mu(\cdot)=\upsilon^{-1}(\cdot)$, or in full detail,	\vspace{-6pt}
\begin{equation}
	Z_t	\approx \upsilon^{(n)}\big(\mu^{(n)}(Z_t,\eta),\eta\big) \mbox{~~~and~~~}
	Z_t	\approx \mu^{(n)}\big(\upsilon^{(n)}(Z_t,\eta),\eta\big) ~,\vspace{-6pt}
\end{equation}
where $\mu^{(n)}(\cdot)$ denotes the future predictive function $\mu(\cdot)$ is applied $n$ times.

In contrast to building a loss function directly by measuring the success of predictions (which would require comparison with at least another transfer~$e(\cdot)$ into the embedding space), we will instead construct a loss solely {\it within} the embedding space that encourages closure of cycles. Conceptually, this allows for far greater representational flexibility and generalisation in the latent space whilst still enforcing high-level temporal consistency.

Fig.~\ref{fig:predictors}~\textit{(right)} depicts all six basic possible predictive cycles amongst current, future, and past embeddings. Thus, our target is to find a space wherein cycle predictions can be made such as to minimise the distance between the start and end of all closed cycles. Thus, 
we arrive at the CEP minimisation objective\vspace{-4pt}
\begin{equation}\label{eq_ccloss}
	\begin{split}
		\mathcal{L}_{C}=
		& ||Z_t-\upsilon(\mu(Z_t,\eta),\eta)||_2 + ||Z_t-\mu(\upsilon(Z_t,\eta),\eta)||_2 \\
		+ & \sum_{n=1}^2||Z_p-\upsilon^{(n)}(\mu^{(n)}(Z_p,\eta),\eta)||_2
		~+ \sum_{n=1}^2||Z_f-\mu^{(n)}(\upsilon^{(n)}(Z_f,\eta),\eta)||_2 ~.\vspace{-12pt}
	\end{split}
\end{equation}
Essentially, this loss sums over the distance errors occurring across the six different basic cycles.
Based on this loss, the set of parameters~$\theta$, that is the union over~$\theta_e$ for the encoder network together with $\theta_{\mu}$ and $\theta_{\upsilon}$, are updated by optimising $\text{arg}\min_{\theta} \mathcal{L}_{C}$. To compute the parameters of this
in practice, the feature encoder and the predictors are co-optimised and evolve together. This process essentially computes an embedding function~$e(\cdot)$ that produces a space where closed cycle prediction is maximally achieved. In tandem with this process, both~$\mu(\cdot)$ and~$\upsilon(\cdot)$ are constructed as actual implementations of cycle prediction.


{\bf Contrastive Loss --} In addition to encouraging cycle consistency, we want the elements~$Z_t$ of the latent space be organised in such a way that distance separates unrelated features, whilst keeping related ones close. In order to implement this, we  introduce a second loss that enforces contrastive feature separation after performing our predictive tasks. 
For this, we adopt InfoNCE \cite{gutmann2010noise}.
In the forward pass, the ground truth representation~$Z_t$ and the predicted representation~$\hat{Z}_t$ are computed as shown in Fig.~\ref{fig:overview}.  $Z_t$ is then treated as the anchor, whilst any
$\hat{Z}_t$ is treated as a positive sample.
Negative samples $\mathcal{I}$ are
taken from mini-batch and memory bank. InfoNCE demands the similarity between the ground truth and positive to be higher than the one between ground truth and negatives. If the similarity is measured by a bi-linear function, a normalised probability~$P_{pos}$ can be represented as\vspace{-6pt}
\begin{equation}\label{eq-cl1}
	P_{pos}=\frac{\exp (S_p/\xi)}{\exp (S_p/\xi) + \sum_{i^* \in \mathcal{I}}\exp (S_{i^*}/\xi)} ~,\vspace{-6pt}
\end{equation}
where $S_p \in \mathbb{R}^{1}$ denotes the similarity result between the positive pair after the bi-linear operation $S_p=Z_t \cdot \hat{Z}_t^T$, $S_{i^*}$ represents the results for negative pairs, respectively, and $\xi$ is the temperature hyper-parameter\cite{wu2018unsupervised}.
Based on this normalised positive pair distribution~$P_{pos}$, given the distribution $\mathcal{T}$, the optimisation task for InfoNCE can be stated as\vspace{-6pt}
\begin{equation}\label{eq-cl}
	\text{arg}\min_{\theta} \mathcal{L}_{X}= -\mathbb{E}_{p\in \mathcal{T}}\bigg(\log P_{pos}\bigg).\vspace{-6pt}
\end{equation}

{\bf Memory Bank --} Contrastive learning benefits from large mini-batch sizes~\cite{he2020momentum}, but given limited GPU memory, it is hard to accommodate both model size and batch size within hardware constraints. Inspired by~\cite{he2020momentum}, we increase the cardinality of possible negative samples~$\mathcal{I}$ by giving up back-propagation of the gradient for negatives. Instead, we use a proxy encoder~$e'(\cdot)$ to generate negative features. The encoder~$e'(\cdot)$ is updated based on the encoder~$e(\cdot)$, but with momentum, {$\theta_{e'} v\leftarrow m\theta_{e'}+(1-m)\theta_e$, given $\theta_e \leftarrow \theta_e+\lambda \nabla \mathcal{L}$,
}
where~$\lambda$ is the learning rate for the encoder~$e(\cdot)$ and~$m \in [0,1)$ is the momentum coefficient.

\begin{figure}[t]
\centering
\includegraphics[width=0.3\linewidth,height=60pt]{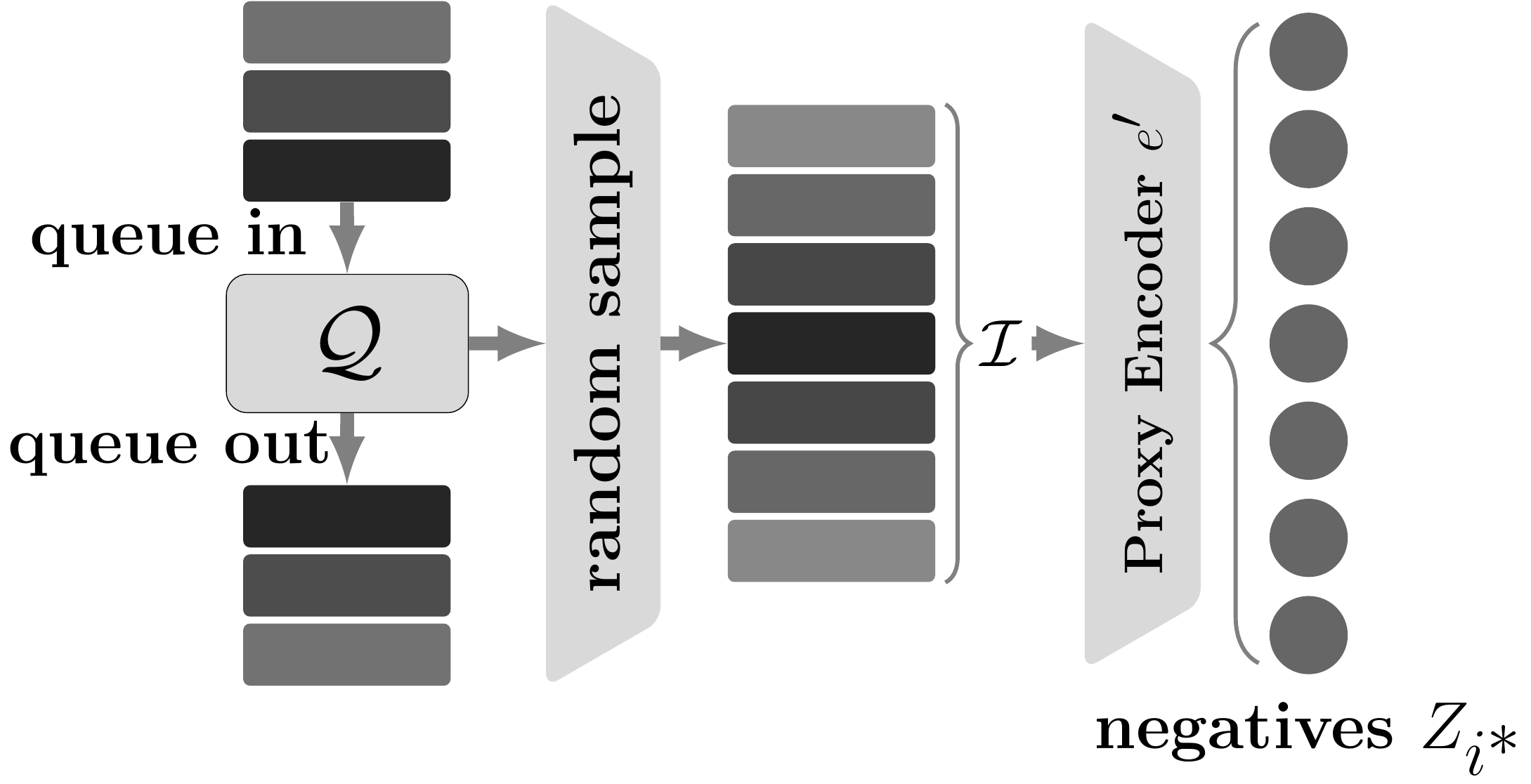}  \hspace{5mm}
	\includegraphics[width=0.24\linewidth,height=60pt]{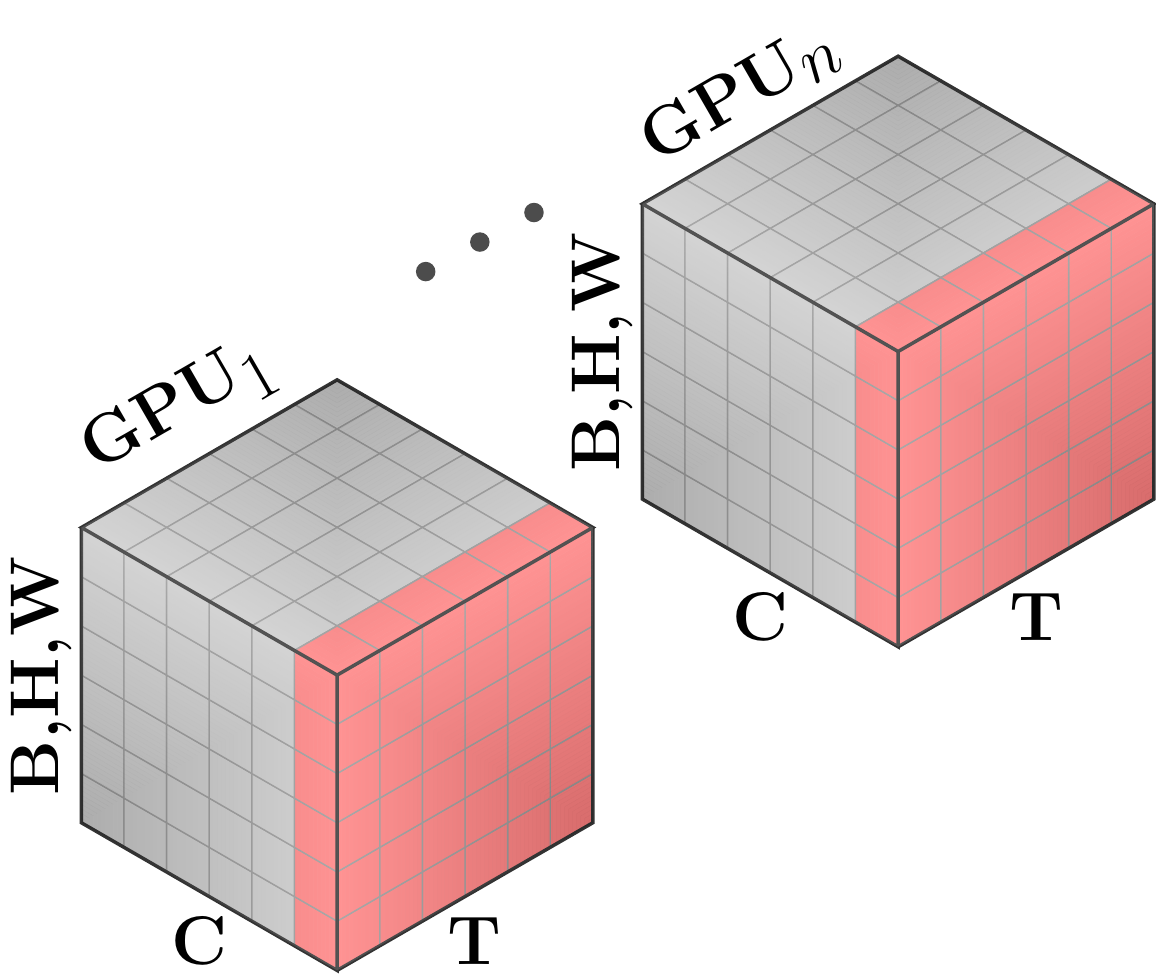}
	\includegraphics[width=0.24\linewidth,height=60pt]{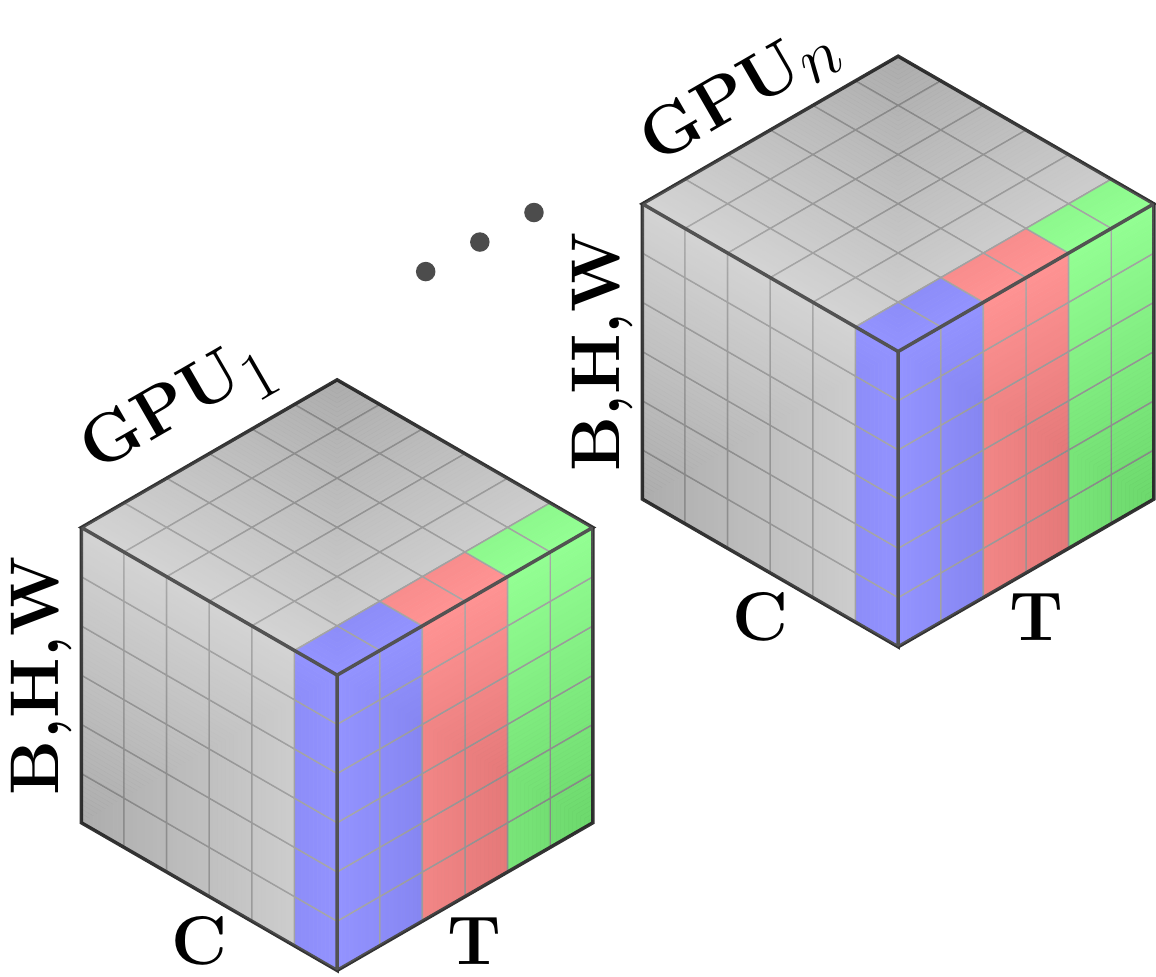}\vspace{10pt}
     \caption{\footnotesize{\textbf{\textit{(left)} Memory Bank.} During each training iteration,~$\mathcal{I}$ samples are picked from a memory queue~$\mathcal{Q}$ to stand-in as source material for negative generation. Following that, the negative features are generated by proxy encoder~$e'(\cdot)$.\textbf{\textit{(middle+right)} Sync-BN vs. Sync-TGN.} In contrast to Sync-BN shown in the middle, Sync-TGN shown on the right organises the temporal dimension into several chunks (purple, pink and green) representing past, present, future snippets in CEP and computes mean and stardard-deviation along the ({\bf B}atch,{\bf H}eight,{\bf W}idth) axes. }}\vspace{-10pt}
     \label{fig:ibn}
\end{figure}

As depicted in Fig.~\ref{fig:ibn} \textit{(left)},~$\mathcal{I}$ samples are picked from memory queue~$\mathcal{Q}$ 
at each iteration. Negative features are then {generated by proxy encoder~$e'(\cdot)$, i.e. $Z_{i^*} = e'(X_{i^*}) ~|~ i^* \in \mathcal{I} \sim \mathcal{Q}$.}

\textbf{Batch Normalization Leaks}. Our encoder~$e(.)$ contains Synchronized Batch Normalization~(Sync-BN) layers. However, intra-temporal communication (\textit{e.g.} Sync-BN) among frame snippets may leak information, such that predictors can learn shortcut solutions. To resolve this, we introduce Synchronized Temporal-Group Normalization~(Sync-TGN) as shown in Fig.~\ref{fig:ibn} \textit{(right)} and ablated
in Section~\ref{sec:ablation} to dilute any leaking signal.

\textbf{Identity Solution}. We note that, theoretically, the identity for both $\mu$ and $\upsilon$ leads to a valid solution for $\mathcal{L}_{C}$~(see Eq.~\ref{eq_ccloss}). Yet, practically we find that both the use of white noise~$\eta$~(see Fig.~\ref{fig:predictors}~\textit{(left)}) and co-optimisation by~$\mathcal{L}_{X}$ disperse mappings in latent space and avoid collapse to an identity cycle. The latter  was never observed in any of our experiments.

\textbf{Optical Flow Obfuscation}. Another possible trivial solution of CEP is the generation of temporal predictions purely from low-level optical flow rather than semantic information. As a counter-measure, we obfuscate optical flow by applying an augmentation strategy to the video snippets,~\textit{e.g.} random horizontal flipping, colour jittering, and random cropping. Note that this preserves semantic temporal coherence. Our direct ablation in Section~\ref{sec:ablation} shows the effectiveness of this approach.

\vspace{-10pt}
\section{Experiments}
\vspace{-5pt}

We now detail the datasets and environments used for learning, and evaluate the effect of CEP pre-training on the downstream task of action recognition. We provide detailed ablation studies to support our design decisions.
\begin{table*}[ht]
	\centering
	\scriptsize
	\begin{tabular}{@{}ll|ccl|ccc@{}}
		\toprule
		\multirow{2}{*}{\textbf{Method}} & \multirow{2}{*}{\textbf{Year}} & \multicolumn{3}{c|}{\textbf{Self-Supervision Training}} & \multicolumn{2}{c}{\textbf{Evaluation}} \\
	   &      & \textbf{Dataset}      & \textbf{Resolution}      & \textbf{Architecture}   & \textbf{UCF101}        & \textbf{HMDB51} \\

		\midrule \midrule
		01 | Random Init.                     &      &        --      & $128\times 128$ & SlowFast     & 56.2          & 23.1  \\
		02 | CEP                     &      & Kinetics-400 & $128\times 128$ & SlowFast          & \textbf{{68.5}}         & \textbf{{34.7}}  \\
		\midrule

		03 | VCOP \cite{xu2019self}            & 2019 & UCF101       & $224\times 224$ & (2+1)D-ResNet       & 72.4          & 30.9  \\
		04 | PRP$^\dagger$ \cite{yao2020video}          & 2020 & UCF101       & $224\times 224$ & (2+1)D-ResNet       & 72.7          & 35.9  \\
		05 | CEP                     &      & UCF101       & $224\times 224$ & (2+1)D-ResNet        & {\textbf{75.5}}         & {36.3} \\
		{06 | PRP + CEP }                  &      & UCF101       & $224\times 224$ & (2+1)D-ResNet        & 73.8         & {\textbf{37.1}}  \\
		\midrule

		07 | DPC \cite{han2019video}           & 2019 & Kinetics-400 & $224\times 224$ & 3D-ResNet34         & 75.7          & 35.7  \\
		08 | Mem-DPC \cite{Han20memdpc}         & 2020 & Kinetics-400 & $224\times 224$ & 3D-ResNet34        &\textbf{ 78.1}          & \textbf{41.2}  \\
		09 | CEP                     &      & Kinetics-400 & $224\times 224$ & 3D-ResNet34             & {76.4}         & {36.5}  \\
        10 | DPC+CEP            &      & Kinetics-400 & $224\times 224$ & 3D-ResNet34         & {\bf 78.1}        & {38.4}  \\

		\midrule
		11 | 3D-Puzzle \cite{kim2019self}      & 2019 & Kinetics-400 & $224\times 224$ & 3D-ResNet18          & 65.8          & 33.7  \\
		12 | RSPNet \cite{chen2020rspnet}                            & 2021 & Kinetics-400 & $224\times 224$ & 3D-ResNet18         & {74.3}          & \textbf{41.8}  \\
		13 | CEP                     &      & Kinetics-400 & $224\times 224$ & 3D-ResNet18            & {\textbf{75.9}}         & {{36.6}}  \\

	    \midrule
    	14 | RSPNet \cite{chen2020rspnet}                            & 2021 & Kinetics-400 & $224\times 224$ & (2+1)D-ResNet         &\textbf{ 81.1}          & \textbf{44.6}  \\
		15 | TCGL \cite{liu2021temporal}                            & 2021 & Kinetics-400 & $224\times 224$ & (2+1)D-ResNet         & 77.6          & 39.7  \\

		16 | CEP                     &      & Kinetics-400 & $224\times 224$ & (2+1)D-ResNet        & 76.7         & 37.6  \\

		\midrule
		17 | CoCLR(RGB) \cite{han2020self}                          & 2020 & Kinetics-400 & $128\times 128$ & S3D-G         & 87.9          & {54.6}  \\
		18 | SpeedNet \cite{benaim2020speednet}                           & 2020 & Kinetics-400 & $224\times 224$ & S3D-G         & 81.1          & {48.8}  \\

	    19 | RSPNet$^\dagger$ \cite{chen2020rspnet}                            & 2021 & Kinetics-400 & $224\times 224$ & S3D-G      &  88.3          & 59.0  \\
		20 | CEP                     &      & Kinetics-400 & $224\times 224$ & S3D-G  & {84.1}         & 46.3  \\
		{21 | RSPNet + CEP}           &      & Kinetics-400 & $224\times 224$ & S3D-G  & \textbf{{90.1}}         & \textbf{59.5 } \\
		\midrule
		22 | CVRL \cite{qian2020spatiotemporal}                           & 2020 & Kinetics-400 & $224\times 224$ & R3D-50($\times$4)          & 92.2          & 67.9  \\
		{23} | {$\rho$BYOL} \cite{feichtenhofer2021large}                           & 2021 & Kinetics-400 & $224\times 224$ & R3D-50($\times$4)          & \textbf{95.5}          & \textbf{73.6}  \\
		\midrule

	\end{tabular}\vspace{7pt}
	\caption{\footnotesize\textbf{Performance Comparison with State-of-the-Art.} The top-1 accuracy rate is reported on downstream action recognition tasks for UCF101 and HMDB51 {in RGB data} in split-1 val split.
	All methods are end-to-end, fine-tuned on the target evaluation set after self-supervised pre-training. $^\dagger$ represents our re-implementation experiments under our directly comparable hardware settings.} \vspace{-12pt}
	\label{tab:result}
\end{table*}


{\bf Datasets -- } For the initial pretraining dataset,  \textbf{Kinetics}~\cite{carreira2017quo} is used in its unlabelled form. It contains 400 action classes, with at least 400 clips per action. Each video lasts around 10s and is taken from YouTube. The actions are human-focused and range from human-object interactions, such as playing instruments, to human-human interactions, such as shaking hands.
\textbf{UCF101}~\cite{soomro2012ucf101} is selected as the primary downstream action recognition task. It contains 13K YouTube videos of different action categories and has a large variety of different camera motions, object appearances, poses and scales, as well as background clutter.
The popular \textbf{HMDB51}~\cite{Kuehne11} is our secondary evaluation dataset with 51 action categories covering a wide range of activities from facial actions to body movements. HMDB51 is often considered more difficult than UCF101 as many classes within it are quite similar.



\textbf{Architecture --} The non-linear encoding function~$e(\cdot)$ is implemented using popular spatio-temporal representation backbones (with  modified last FC layer yielding an output feature size of 2048,
\textit{e.g.} SlowFast~\cite{feichtenhofer2019slowfast}, (2+1)D-ResNet \cite{tran2018closer}, 3D-ResNet \cite{hara2017learning}, and S3D-G \cite{xie2018rethinking},
allowing us to both compare directly with other methods and to compare these architectures. The two predictors~$\upsilon$ and~$\mu$ are implemented with 4-layers MLP (see Fig.~\ref{fig:predictors}).

\textbf{Pre-processing --} Video snippets~$X_t$ of 16~frames each are formed by sampling the original 30fps video at a rate of 6~fps. Frames are selected with a consistent stride to preserve the regularity of temporal dynamics. We sample 3 snippets without overlap to form associated triples. For augmentation, we apply snippet-wise random horizontal flip and random colour jittering. Finally, snippets are rescaled and centre-cropped to $224\times224$ resolution.

\textbf{Training --} We implemented CEP in Pytorch using distributed training on 8 GPUs, each with a mini-batch size of 3 with input tensor sizes being~$\mathbb{R}^{3\times3\times16\times224\times224}$. {To produce predictor extensions, we use Gaussian noise with~$\eta \in \mathbb{R}^{1024}$.} Similar to~\cite{mnih2015human}, we bootstrap the groundtruth representation $Z_t$ from encoder~$e(.)$ with its weights shifted {to those from 5000 iterations before, resulting in faster convergence.} 

For cycle consistency, we run all 6~possible cycles (see Fig.~\ref{fig:predictors}~{\textit{(right)}}) and use the overall loss~$\mathcal{L_C}$ for back-propagation.
For contrastive learning, the negative features are sampled from the memory bank with the proxy encoder~$e'(\cdot)$ using the momentum coefficient~$m=0.9$ {and InfoNCE temperature $\xi=1$}. We sample 16 easy negative features from the memory bank, and use 4 hard negative features in the mini-batch where easy negatives are from different videos, whilst the hard negatives are from the same video. The ratio between easy and hard samples is 4:1 
so as to make the learning task neither too hard to learn nor too easy to overfit.
All the models are trained end-to-end using SGD as optimiser with an initial learning rate of~$10^{-2}$, momentum~$0.9$ and weight decay~$10^{-4}$. The system requires to minimise the combined objective $ \mathcal{L}=\lambda\mathcal{L}_{C}+\mathcal{L}_{X}$, where $\lambda$ is set to 0.1 intuitively. During inference, video snippets from the validation set are sampled using the same strategy mentioned above, but all the augmentations are removed maintaining a scaled centre-crop.

\textbf{Compute Requirements --} Quantifying CEP's computational footprint with different backbone architectures shows the approach has a very light parameter impact yet significant FLOPS cost. For the S3D-G backbone, the CEP concept increases parameters from $11.04M$ to $13.16M$ and FLOPS from $34.3G$ to $85.6G$. For the R(2+1)D backbone, CEP increases parameters from $15.02M$ to $17.15M$ and FLOPS from $171.8G$ to $429.5G$.  For 3D-ResNet-34, CEP increases parameters from $64.17M$ to $66.29M$ and FLOPS from $100.58G$ to  $261.06G$.
{For example, using 3D-ResNet18 for CEP training~(see Table~\ref{tab:result}, row 13) takes approx.~1 week for 50 epochs with 8 NVIDIA P100 GPUs running on Lenovo nx360 nodes with 2.4 GHz Intel E5-2680 v4 (Broadwell) CPUs and 128 GB of RAM using the Kinetics-400 dataset as data source with batch size~24.}


\vspace{-10pt}
\section{Results}
\vspace{-5pt}
We compare CEP's performance by looking at different state-of-the-art self-supervised methods, fine-tuning on the action classification tasks of UCF101 and HMDB51. Our results are reported based on split-1 validation accuracy.

\textbf{Baselines --} At the lower resolution of $128\times128$ (rows {01--02} in Table~\ref{tab:result}), we note the positive effect of CEP pre-training against a random initialisation with a significant boost of 12.3\% on UCF101 and 11.6\% on HMDB51.
Next,  training only on the UCF101 dataset at $224\times224$ resolution and on a (2+1)D-ResNet (\textit{i.e.}~{rows 03--05}),  our CEP approach outperforms other comparable methods, such as VCOP \cite{xu2019self} and PRP \cite{yao2020video}, reaching {75.5\% and 36.3\%} accuracy for UCF101 and HMDB51, respectively.
Methods 07--10 show a direct comparison with DPC~\cite{han2019video}. Given the same 3D-ResNet34 backbone, {CEP exceeds DPC \cite{han2019video} at 76.4\% and 36.5\% on UCF101 and HMDB51, respectively. When we integrate the CEP concept into  DPC (to get DPC+CEP), a further improvement leading to 78.1\% and 38.4\% can be observed on those two datasets, respectively. UCF101 MemDPC~\cite{Han20memdpc} results are on-par with DPC+CEP {which uses fewer parameters and $10\%$ fewer FLOPS}}. On rows 11--20 in Table \ref{tab:result}, depending on the backbone network, CEP exhibits a mixed performance compared to other recent networks. In particular, with a 3DResNet18 backbone CEP outperforms 3D-Puzzle \cite{kim2019self} by a significant margin on both datasets, while also outperforming RSPNet \cite{chen2020rspnet} on UCF101.
On the (2+1)D-ResNet backbone,  CEP on its own achieves competitive performance in comparison to TCGL \cite{liu2021temporal}. RSPNet on the other hand outperforms both CEP and TCGL in this category. When fuelled by a S3D-G backbone, RSPNet dominates CEP, yet CEP exceeds Google's SpeedNet~\cite{benaim2020speednet} on UCF101, while being competitive on HMDB51.

{\textbf{Add-on Efficacy of CEP --} However, we find that adding CEP to leading architectures~(rows 06, 10, 21) while keeping their main concepts unchanged consistently outperforms their original benchmarks supporting general efficacy of our concept. Specifically, to use CEP as an add-on we introduce forward and backward prediction modules on top of feature embeddings of other architectures. We then contrastively learn the predictive embeddings and enforce cycle consistency by introducing our $\mathcal{L}_X$ and $\mathcal{L}_C$ losses, respectively, to the overall loss. For PRP with a (2+1)D-ResNet backbone, integrating the CEP concept can introduce a 1.1\% and 1.2\% performance boost for UCF101 and HMDB51, respectively~(\textit{i.e.}~{rows 04 vs. 06}). For DPC with a 3D-ResNet34 backbone, DPC+CEP increases the baseline by 2.4\% and 2.7\%, respectively~(\textit{i.e.}~{rows 07 vs. 10}). The state-of-the-art RSPNet can also benefit from CEP with an improved benchmark for RSPNet+CEP of 90.1\% on UCF101 and 59.5\% on HMDB51~(\textit{i.e.}~{rows 19 vs. 21}). For the architectures tested so far this produces a leading performance~({rows 1--21}). We conclude that CEP is consistently and demonstrably effective as an add-on technique across the architectures tested. }

\textbf{Massively Large Setups --} CVRL \cite{qian2020spatiotemporal} on row 22 {and $\rho$BYOL \cite{feichtenhofer2021large} on row 23} show impressive performance on UCF101 and HMDB51. Yet, these works are not directly apple-to-apple comparable since setups run at massively larger scale. Particularly, huge batch sizes (CVRL: 1024 vs. CEP: 24 ), very large backbone networks (CVRL: 31.7M vs. rhoBYOL: 31.8M vs. CEP: 13.06M parameters) and more than an order of magnitude further pre-train epochs (CVRL: 800 vs. rhoBYOL: 800 vs. CEP: 50 epochs) are used. We are unable to offer direct comparisons to these massive experiments as we are limited by computational resources.
{
Nevertheless, our results showed that the core CEP concept effectively promotes semantic learning, is valuable and widely applicable as an add-on, where it enhanced the performance of all X+CEP architectures tested.}


\begin{table}[t]
\begin{minipage}[t]{0.3\textwidth}
\begin{center}
\textbf{\footnotesize\textit{(a)}}\\
\scriptsize
\begin{tabular}[t]{c|c}
	\multirow{2}{*}{\begin{tabular}[c]{@{}c@{}}\textbf{{Pre-Training}}\\ \textbf{{Loss Functions}}\end{tabular}} 	   &
	    \multirow{2}{*}{\begin{tabular}[c]{@{}c@{}}\textbf{Eval. (UCF)}\\ \textbf{Top1 acc}\end{tabular}} \\
        \\ \hline
	    random init.             & 56.2   \\
		 $\mathcal{L}_{C}$ only,~($ \mu~and \upsilon $) & 61.1\\
		$\mathcal{L}_{X}$ only,~($\mu=\upsilon$) & 63.7\\
		$\mathcal{L}_{X}$ only,~($ \mu~and \upsilon $) & 64.6\\
	$\mathcal{L}_{X}$+$\mathcal{L}_{C}$,~($ \mu~and \upsilon $)                      & \textbf{68.5} \\
	\end{tabular}
\end{center}
\end{minipage}
\begin{minipage}[t]{0.3\textwidth}
\begin{center}
\textbf{\footnotesize\textit{(b)}}\\
\scriptsize
\begin{tabular}[t]{c|c}
	\multicolumn{1}{c|}{\textbf{Optical Flow}} &
	 \multirow{2}{*}{\begin{tabular}[c]{@{}c@{}}\textbf{Eval. (UCF)}\\ \textbf{Top1 Acc}\end{tabular}} \\

       \textbf{Obfuscation} &      \\
       \hline
	 \xmark                   & 70.3 \\
	 \cmark                   & \textbf{75.4 }\\
	\end{tabular}
	\end{center}
\end{minipage}
\begin{minipage}[t]{0.3\textwidth}
\vspace{-6pt}
\begin{center}
\textbf{\footnotesize\textit{(c)}}\\
\scriptsize
	\begin{tabular}[t]{c|c|c|c}
		 \multicolumn{3}{c|}{\textbf{Memory Bank Strategies}} & \multirow{2}{*}{\begin{tabular}[c]{@{}c@{}}\textbf{Eval. (UCF)}\\ \textbf{Top1 Acc}\end{tabular}} \\

	\textbf{M-Bank} & \textbf{Strategy} & $\mathbf{m}$ & \\ \hline
		 \xmark  & -        & -   & 71.2\\
	\cmark  & static   & -   & 72.8\\
 \cmark  & dynamic  & 0.1 & 73.3 \\
	    \cmark  & dynamic  & 0.9 & \textbf{75.4}\\
	\end{tabular}
	\end{center}
\end{minipage}\\ \ \\
\begin{minipage}[t]{0.28\textwidth}
\begin{center}
\textbf{\footnotesize\textit{(d)}}\\
\scriptsize
	\begin{tabular}[t]{c|c}
	 \multicolumn{1}{c|}{\textbf{Normalisation}} &
	 \multirow{2}{*}{\begin{tabular}[c]{@{}c@{}}\textbf{Eval. (UCF)}\\ \textbf{Top1 Acc}\end{tabular}} \\
	    \textbf{Strategies}& \\
	    \hline
 Sync-BN          & 71.8 \\
 Sync-TGN           & \textbf{75.4 }\\
	\end{tabular}
	\end{center}
\end{minipage}
\begin{minipage}[t]{0.31\textwidth}
\begin{center}
\textbf{\footnotesize\textit{(e)}}\\
\scriptsize
	\begin{tabular}[t]{c|c}
	\multirow{2}{*}{\begin{tabular}[c]{@{}c@{}}\textbf{Temporal}\\ \textbf{Receptive Field}\end{tabular}} 	   &
	   \multirow{2}{*}{\begin{tabular}[c]{@{}c@{}}\textbf{Eval. (UCF)}\\ \textbf{Top1 acc}\end{tabular}} \\
		\\ \hline
	 1.5s        & 73.7 \\
 3s          & {\bf 75.4} \\
	\end{tabular}
	\end{center}
\end{minipage}
\begin{minipage}[t]{0.3\textwidth}
\vspace{-6pt}
\begin{center}
\textbf{\footnotesize\textit{(f)}}\\
\scriptsize
	\begin{tabular}[t]{c|c|c}
		 \multicolumn{2}{c|}{\textbf{Feature Dimensionality}} & \multirow{2}{*}{\begin{tabular}[c]{@{}c@{}}\textbf{Eval. (UCF)}\\ \textbf{Top1 Acc}\end{tabular}} \\
\textbf{temp. pres.} & \textbf{output size }& \\
\hline

 \cmark                 & $2\times2048$               & 73.3  \\
 \xmark                 &  $2048$                     & 71.2 \\
\xmark                 &  $4096$                     & \textbf{75.4}  \\
	\end{tabular}
	\end{center}
\end{minipage}\vspace{3pt}
\caption{\footnotesize\textbf{Detailed Ablation Studies.} Verification of key component effectiveness via ablation. All studies pre-train on Kinetics-400 with a SlowFast Resnet50 encoder and evaluate performance on UCF101. All experiments apart from \textit{(a)} use a resolution of~$224\times 224$ for experiments. Note that all design choices demonstrably impact positively on performance. {In particular, as shown in \textbf{\textit{(a)}}, the use of independent forward and backward predictors~$ \mu~and \upsilon $ has value and improves performance, as does the addition of each of the proposed core losses $\mathcal{L}_{X}$ and~$\mathcal{L}_{C}$.  For instance, adding cycle consistency~$\mathcal{L}_{C}$ to an otherwise fixed setup is effective adding $3.9\%$ to accuracy. }}\vspace{-15pt}
\label{tab:ablation}
\end{table}

\vspace{-12pt}
\section{Ablation Experiments} \label{sec:ablation}
\vspace{-7pt}
{In this section, we present a detailed ablation study in order to quantify the impact of the CEP concept and validate the effectiveness of the components in the basic setup. In particular, we will show that cycle consistency $\mathcal{L}_{C}$ improves learning as it encourages an inverse relationship between otherwise independent forward and backward predictors $\mu(\cdot)$ and $\upsilon(\cdot)$.}

{{\bf Efficacy of both Contrastive and Cycle Loss -- }
For experiments that quantify the importance of contrastive loss $\mathcal{L}_{X}$ and cycle loss $\mathcal{L}_{C}$ during training, we use a fixed SlowFast setup with a ResNet50 as the encoder. The input is rescaled and randomly cropped to a resolution of $128\times128$. The models are pre-trained on the Kinetics-400 dataset as before, and results are reported after fine-tuning on UCF101 dataset for 30~epochs. As shown in Table~\ref{tab:ablation}~\textit{(a)}, both $\mathcal{L}_{X}$  and  $\mathcal{L}_{C}$ contribute significantly to CEP's performance, which drops considerably when removing any one of the losses. The addition of cycle consistency
to an otherwise fixed system improves accuracy significantly by~$3.9\%$.}

{
{\bf Consistency of $\mathcal{L}_{C}$ Efficacy -- }  To show coherent efficacy of the cycle consistency concept throughout the training process, we compare the full CEP setup against one that does not use $\mathcal{L}_{C}$ at different points of the pre-training process -- the same comparison as shown after full pre-training in the last two lines of Table~\ref{tab:ablation}~\textit{(a)}. Table~\ref{tab:steps} shows the results confirming that throughout pre-training the use of $\mathcal{L}_{C}$ is beneficial and able to enhance action recognition.}

\begin{table}[t]
\begin{center}
\footnotesize
\begin{tabular}{l|ccc}
\textbf{{Pre-Training Loss Functions}} & 179K-step & 358K-step & 537k-step \\
\hline
	No Cycle Consistency used: $\mathcal{L}_{X}$ only & 61.0      & 64.2      & 64.6      \\
    Cycle Consistency used: \textcolor{white}{......}$\mathcal{L}_{X}$+$\mathcal{L}_{C}$ & \textbf{63.1}      & \textbf{64.8}      & \textbf{68.5}
\end{tabular}
\end{center}
\caption{{\footnotesize\textbf{Consistency of $\mathcal{L}_{C}$ Efficacy throughout Pre-Training.} UCF101 Top 1 accuracy performance when pre-training for different numbers of steps with and without cycle consistency $\mathcal{L}_{C}$, fixing all other settings. The results confirm that the use of cycle consistency consistently benefits performance.}}\vspace{-15pt}
\label{tab:steps}
\end{table}

{\bf Obfuscation of Inter-clip Optical Flow -- }   
We will now ablate further components of the basic setup beyond the core concepts. As discussed earlier and quantified in Table~\ref{tab:ablation}~\textit{(b)}, the augmentation-based obfuscation of inter-clip optical flow does indeed improve learning and boosts accuracy significantly by~$5.1\%$ when tested at~$224\times 224$ resolution.

{\bf Memory Bank Strategies --}
As shown in Table~\ref{tab:ablation}~\textit{(c)}, when tested again at~$224\times 224$ resolution learning benefits from a large batch size as demonstrated recently for other contrastive learning in~He~\textit{et al}.~\cite{he2020momentum}. Feature memory banks can provide this property at controllable cost. We compare two memory bank strategies: (i) static negative features are sampled from the memory bank, and (ii) negatives are generated by a proxy encoder~$e'(\cdot)$ dynamically. These models take 4~negatives from the mini-batch and sample 16~negatives from the memory bank per step. We compare these settings against a model without a memory bank that just takes 4 negatives per step from the mini-batch. We conclude that a dynamic memory bank with a large momentum factor~$m$ can clearly provide benefits to CEP learning.

{\bf Synchronized Temporal-Group Normalization -- } 
Next, we explore the effect of our Sync-TGN in preventing the CEP spiralling into a trivial solution.
For the BN experiment, we use the same batch size (3 per GPU) for both sync-BN and sync-TGN, and we have 8 GPUs in both scenarios. Table~\ref{tab:ablation}~\textit{(d)} depicts the effectiveness of Sync-TGN over Sync-BN showing an increase of 3.6\% in self-supervised learning accuracy, 
supporting our hypothesis
that information leak can be mitigated to some extent.

%

{\bf Temporal Receptive Field -- }  
To investigate the effect of the temporal receptive field~(RF), we sparsely sample 16 frames as the input. These 16 frames cover  either 1.5s or 3s of the input video. As Table~\ref{tab:ablation}~\textit{(e)} shows,
 the input clip with the larger temporal RF increases the
 learning performance by 1.7\%, which suggests that the model benefits from longer exposure to scene
 dynamics 
 for learning of semantics.

{\bf Feature Space Dimensionality -- }
Finally, Table~\ref{tab:ablation}~\textit{(f)} summarises our evaluations on  the influence of changing the dimensionality of the embedding space. Unsurprisingly, a larger representational space benefits performance. Note that for noise~$\eta$, we use half of the embedding's feature size, \textit{e.g.} if~$Z_t\in\mathbb{R}^{2048}$ then~$\eta\in\mathbb{R}^{1024}$. Splitting the feature space artificially into sub-spaces to preserve temporal separation was found to be ineffective.

\vspace{-12pt}
\section{Conclusion}
\vspace{-7pt}
In this paper, we showed that action classification can benefit from learning feature spaces of video snippets in which temporal cycles are maximally predictable. We introduced Cycle Encoding Prediction
which can effectively encode video in this latent space using the concepts of closed temporal cycles and contrastive feature separation. CEP can be seen as a temporal regularisation approach used to guide the construction of latent semantic video spaces. It unlocks the related self-supervision signal and is demonstrably effective when used for pretext learning. Ablation studies support the effectiveness of the core loss functions and system components.
We reported results across different backbones for the standard datasets UCF101 and HMDB51.

{We showed that the CEP concept introduces an effective and elegant self-supervision criterion for video that can serve as an additional guide for pretext training. As an add-on concept it was shown to enhance the performance of many existing architectures.}

\bibliography{cep}

\end{document}